\documentclass[letterpaper, 10 pt, conference]{ieeeconf}
\IEEEoverridecommandlockouts
\usepackage{makeidx}         
\usepackage[pdftex]{graphicx}
\usepackage{epstopdf}
\usepackage{multicol}        

\usepackage{hyperref}
\usepackage{amsmath,amssymb}
\usepackage{euscript}
\usepackage{url}
\usepackage{mathrsfs}
\usepackage{array}
\usepackage{wrapfig}
\usepackage{subfigure}

\usepackage{epsfig ,alltt}
\usepackage{psfrag,xr}
\usepackage{pdfpages}
\usepackage{epstopdf}
\usepackage{pstool}
\usepackage[noadjust]{cite}

\usepackage{bm}
\usepackage{multirow}
\usepackage{balance} 

\newcommand{\figref}[1]{Fig.~\ref{#1}}
\newcommand{\secref}[1]{Sec.~\ref{#1}}

\begin{document}

\title{\LARGE \bf Learning Agile Locomotion and Adaptive Behaviors via RL-augmented MPC}

\author{Yiyu Chen and Quan Nguyen
\thanks{This work is supported in part by National Science Foundation Grant IIS-2133091. The opinions expressed are those of the authors and do not necessarily reflect the opinions of the sponsors.}
    \thanks{Y. Chen and Q. Nguyen are with the Department of Aerospace and Mechanical Engineering, University of Southern California, Los Angeles, CA 90089, email: {\tt yiyuc@usc.edu, quann@usc.edu}.}
}
\maketitle
\begin{abstract}
 In the context of legged robots, adaptive behavior involves adaptive balancing and adaptive swing foot reflection. While adaptive balancing counteracts perturbations to the robot, adaptive swing foot reflection helps the robot to navigate intricate terrains without foot entrapment. In this paper, we manage to bring both aspects of adaptive behavior to quadruped locomotion by combining RL and MPC while improving the robustness and agility of blind legged locomotion. This integration leverages MPC's strength in predictive capabilities and RL's adeptness in drawing from past experiences.  Unlike traditional locomotion controls that separate stance foot control and swing foot trajectory, our innovative approach unifies them, addressing their lack of synchronization. At the heart of our contribution is the synthesis of stance foot control with swing foot reflection, improving agility and robustness in locomotion with adaptive behavior. A hallmark of our approach is robust blind stair climbing through swing foot reflection. Moreover, we intentionally designed the learning module as a general plugin for different robot platforms. We trained the policy and implemented our approach on the Unitree A1 robot, achieving impressive results: a peak turn rate of 8.5 rad/s, a peak running speed of 3 m/s, and steering at a speed of 2.5 m/s. Remarkably, this framework also allows the robot to maintain stable locomotion while bearing an unexpected load of 10 kg, or 83\% of its body mass. We further demonstrate the generalizability and robustness of the same policy where it realizes zero-shot transfer to different robot platforms like Go1 and AlienGo robots for load carrying. Code is made available for the use of the research community at \url{https://github.com/DRCL-USC/RL_augmented_MPC.git}
\end{abstract}

\section{Introduction}


In the quest for the practical deployment of quadruped robots in real-world scenarios, the integration of adaptive behavior into their motion remains a challenge. This adaptive behavior consists of two essential dimensions: 1) real-time adaptation to external perturbations, and 2) self-adjustments such as foot reflection when a robot's foot gets stuck in an obstacle. Current advancements in legged mobility predominantly lean on Model Predictive Control (MPC) and Reinforcement Learning (RL). MPC, which employs real-time optimization over a set horizon to compute the optimal control sequence, often requires substantial computational resources and careful parameter tuning. RL, while notable for exceptional adeptness at navigating uneven terrains and unexpected disturbances, demands extensive offline computation, and careful reward tuning, and often produces policies tailored to specific robots. To combine the benefits of both MPC and RL, we present an innovative approach synthesizing the strengths of model-based control and reinforcement learning. Our central objective is to bolster agility, robustness, and adaptive behavior in blind locomotion through the integration of stance foot control and swing foot reflection using RL.
\begin{figure}[t]
    \centering

        \includegraphics[clip, width = \linewidth]{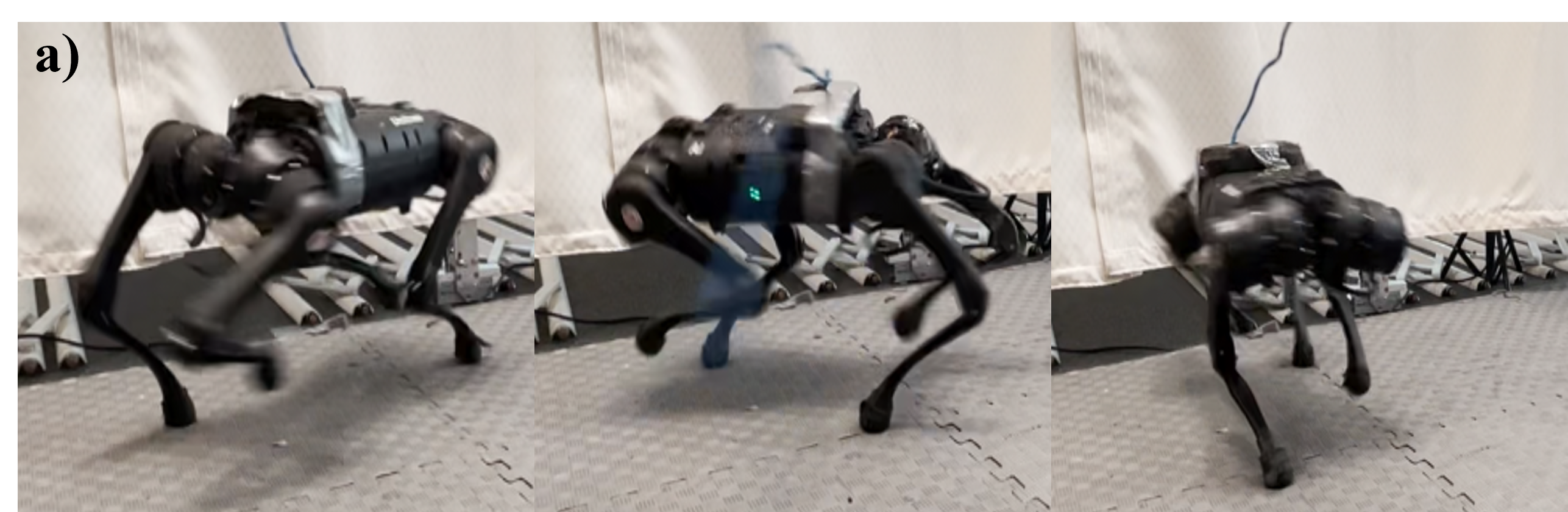} 

        \vspace{0.2em}
        \includegraphics[trim={0cm, 1.5cm, 0.0cm, 0cm}, clip, width = \linewidth]{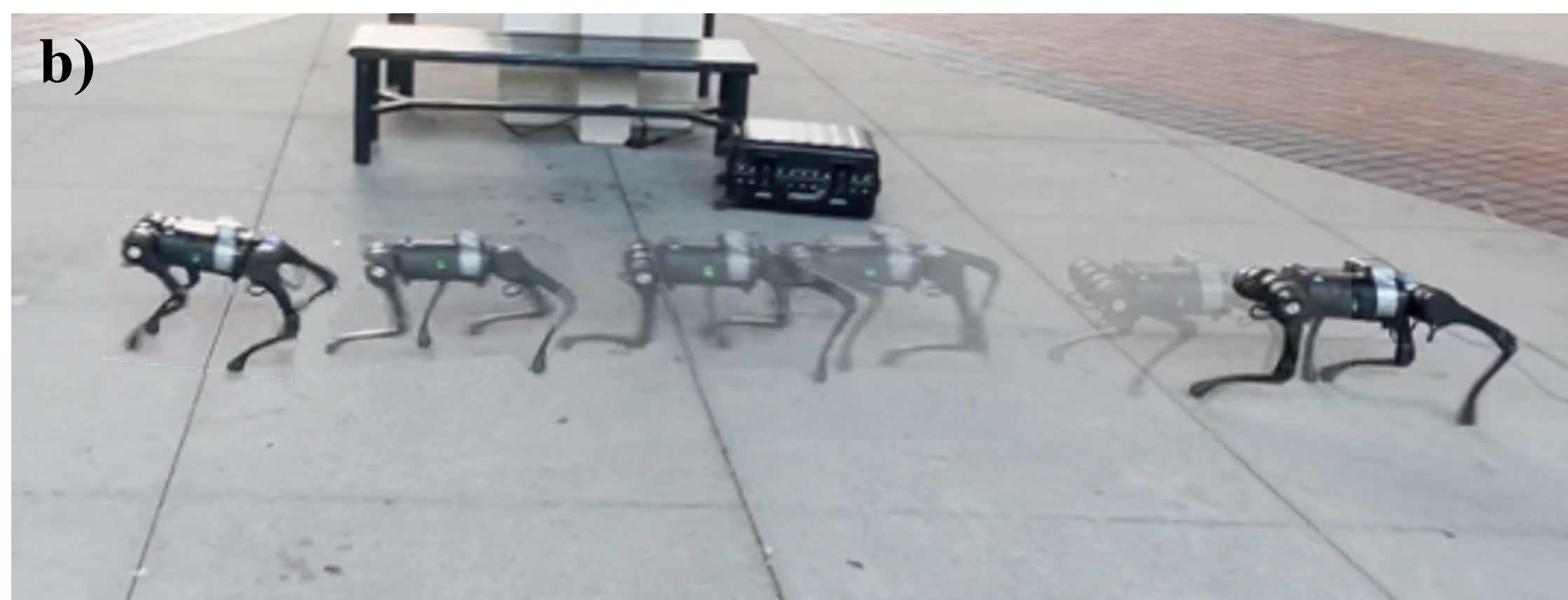}

         \vspace{0.2em}
        \includegraphics[trim={0.0cm, 5cm, 0.3cm, 1cm},clip, width = \linewidth]{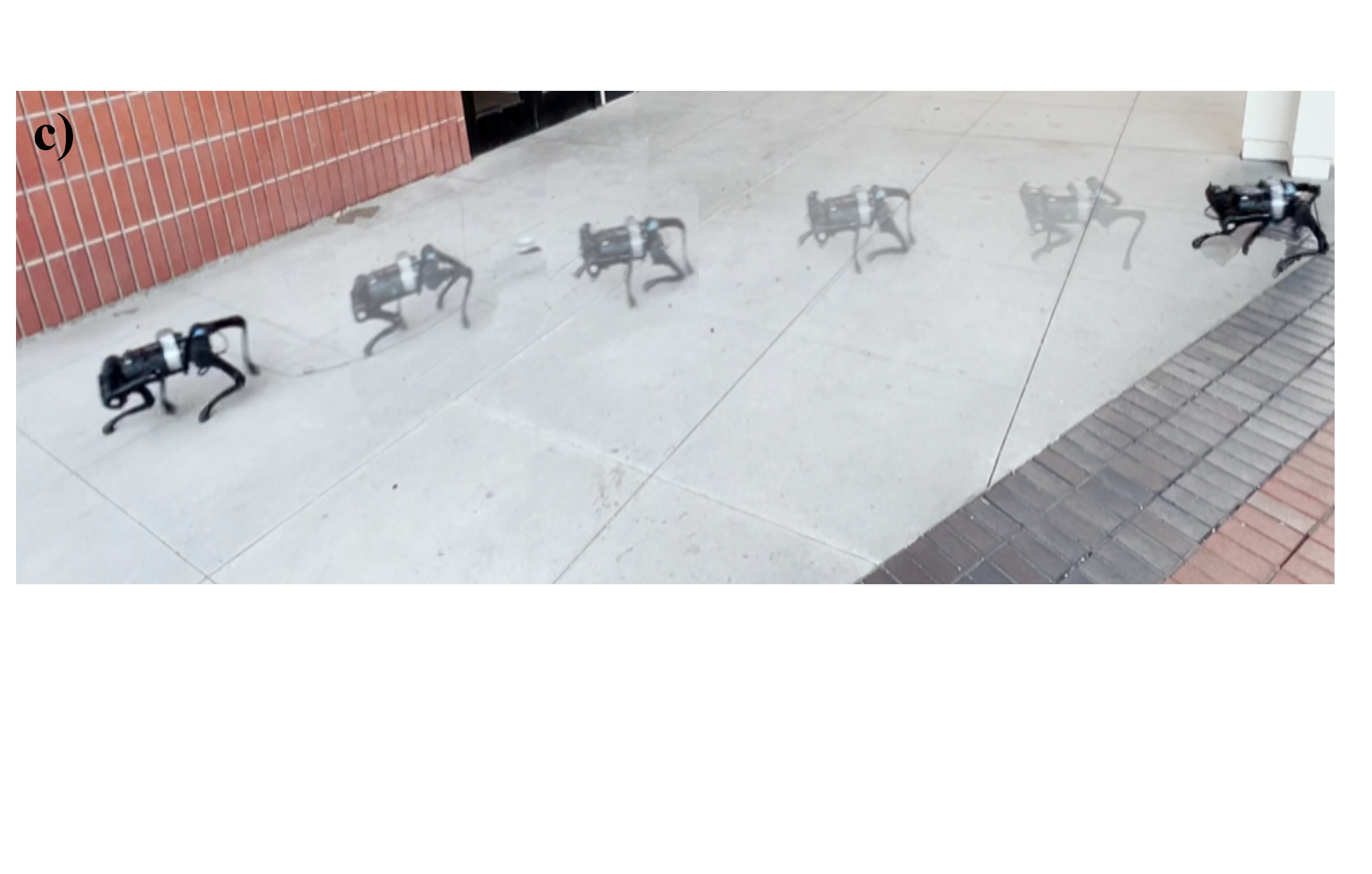}

         \vspace{0.2em}
        \includegraphics[trim={0cm, 0cm, 0cm, 0cm},clip,width = \linewidth]{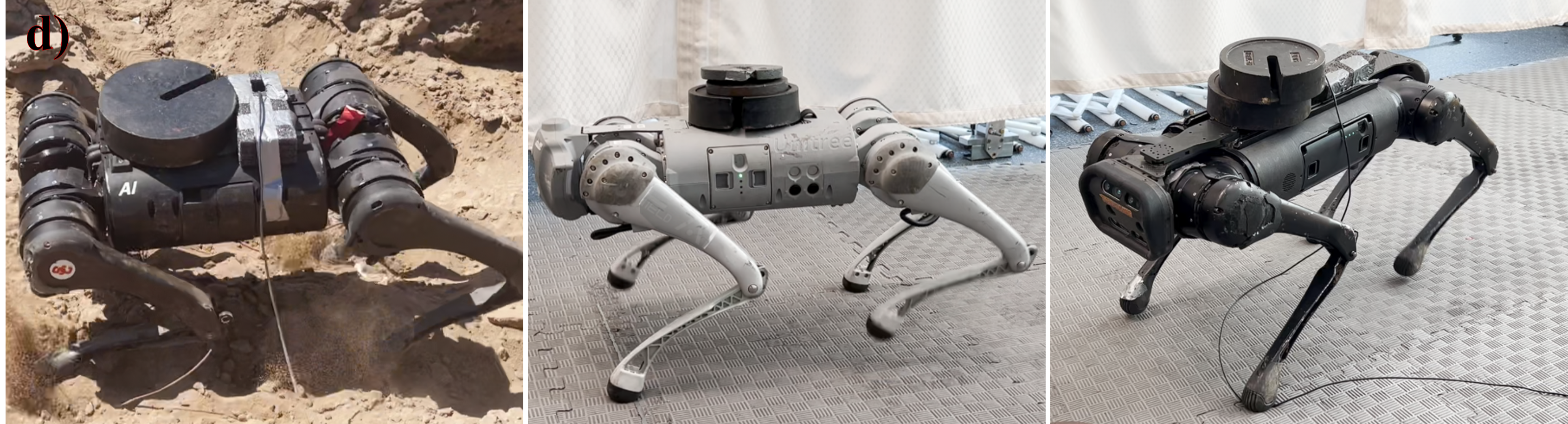}

        \includegraphics[width = \linewidth]{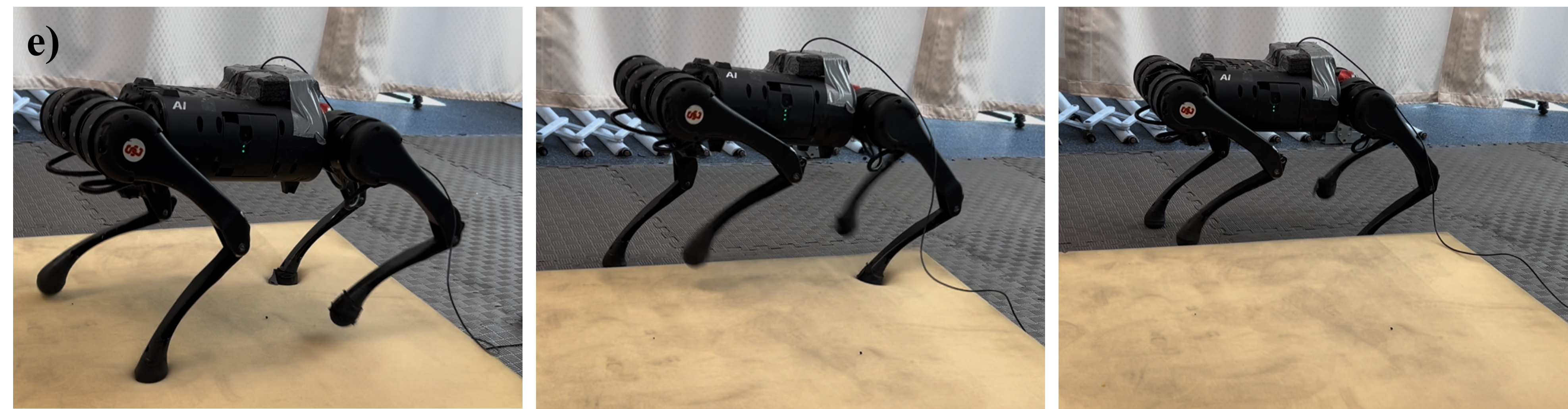}

    \caption{Experiment result highlights. a) High-speed steering in place; b) High-speed running; c) High-speed running and steering; d) Generalization of the same policy across different robot platforms; e) Transition between soft and hard terrain. Experiment video: \url{https://www.youtube.com/watch?v=HxSIxTnEw08}}
    \label{fig:highlight}
    \vspace{-2em}
\end{figure}

MPC, as validated by multiple studies ~\cite{di2018dynamic, kim2019highly, ding2021representation, grandia2023perceptive, chi2022linearization, rathod2021model, sombolestan2023hierarchical} has gained traction in the legged robot community for its capability to handle the hybrid nature of quadrupedal locomotion under constraints. Central to the MPC paradigm is its prediction based on simplified dynamics, offering a future state estimation while preserving real-time computational feasibility. Yet, these simplified models inherently come with model uncertainties. For instance, the single rigid dynamics (SRB) model overlooks the resultant dynamics from leg momentum and external disturbance. Further complexities arise when translating these optimized trajectories into joint-level commands. Existing methodologies adopt a hierarchical control structure for this conversion, using techniques like Jacobian mapping \cite{bledt2018cheetah}, control barrier functions\cite{grandia2021multi}, and joint-level whole body control\cite{farshidian2017efficient, kim2019highly}. The problem of addressing uncertainties in legged motion has been approached with adaptive control methods \cite{cao20081, nguyen20151, nguyen2018dynamic, sreenath2013embedding, minniti2021adaptive, sombolestan2021adaptive, sombolestan2023adaptive}, adjusting the control parameters online. In addition, an implicit limitation of the MPC framework is its inherent decoupling of stance foot control from swing foot control due to the intricate modeling challenges of their interplay. 

Reinforcement learning offers a tantalizing alternative~\cite{lee2020learning, margolis2022rapid, krishna2023learning, makoviychuk2021isaac, ji2022hierarchical, kumar2021rma, cheng2023legs, feng2023genloco, margolis2023walk,jin2022high}, implicitly deciphering the dynamics interplay between the stance and swing control for all kinds of locomotion. In this paradigm, agents continually engage with environments, iteratively refining their action strategies based on the reward, resulting in the mastery of complex terrains and adaptive behavior attuned to environmental dynamics. Nevertheless, deploying RL in real-world scenarios raises legitimate concerns about its generalizability and safety. The aforementioned challenges underscore the urgency for a control framework evolution, one that concurrently addresses model uncertainties and optimizes swing foot reflection with regularized motions. 

Researchers integrate MPC and reinforcement learning to combine the benefits of RL and model-based control. In \cite{kang2023rl+}, the RL framework utilizes model-based optimal control to generate reference motion and then leverages motion imitation technique\cite{peng2020learning} to learn versatile legged motion. RL~\cite{sacks2022learning, yang2022fast, yang2023cajun} is also utilized to learn parameters or dynamics in the MPC problem. \cite{xie2022glide} proposes an RL-based control of the accelerations of an SRB model which allows robust sim-to-real transfer. \cite{pandala2022robust} leverages RL to infer the set of unmodeled dynamics for the RMPC framework for adaptive locomotion. Additionally, \cite{sun2021online} proposed an online supervised learning technique to derive a residue model to address the model uncertainties for model-based controller. What sets our research apart from these studies is our innovative synthesis of stance foot control and swing foot reflection by leveraging RL, enabling adaptive balancing and adaptive foot reflection within one unified framework.

In this paper, we present a novel RL-augmented MPC framework tailored to enhance blind locomotion for quadruped robots. By leveraging RL, we synthesize stance foot control and swing foot reflection from the convex MPC framework \cite{di2018dynamic}, specifically addressing the inherent issues of model uncertainty and the pre-defined swing foot trajectory. By tackling the dual challenges of adapting to model uncertainty and optimizing foot reflection, we successfully demonstrated improved agility, robustness, and adaptive behavior in blind legged locomotion. Notably, our research introduces robot-agnostic action and observation spaces to guarantee the policy's generalizability across various robot platforms. Our proposed framework has the following contributions:
 \begin{itemize}
     \item We introduce a novel RL-augmented MPC framework designed for adaptive blind quadruped locomotion, encompassing high-speed movement, uncertain dynamics adaptation, and reactive obstacle traversal.
     
     \item  Our contribution uniquely combines stance foot force control with swing foot reflection, addressing model uncertainties and bridging foot swing and force control, overcoming inherent challenges in the nominal MPC framework.
     
     \item Our framework provides a robot-agnostic RL module for MPC, realizing zero-shot transfer across various robot platforms, and showcasing state-of-the-art performance on the Unitree A1, Go1, and AlienGo robots.
    
 \end{itemize}

 The paper is organized as follows: \secref{sec:approach} presents our novel RL augmented MPC framework. Then, experimental validation is presented in \secref{sec:result}. \secref{sec:conclusion} provides concluding remarks.

\begin{figure*}[h]
    \centering
    \includegraphics[trim={0.1cm 3.5cm 0.1cm 3.8cm},clip, width = \linewidth]{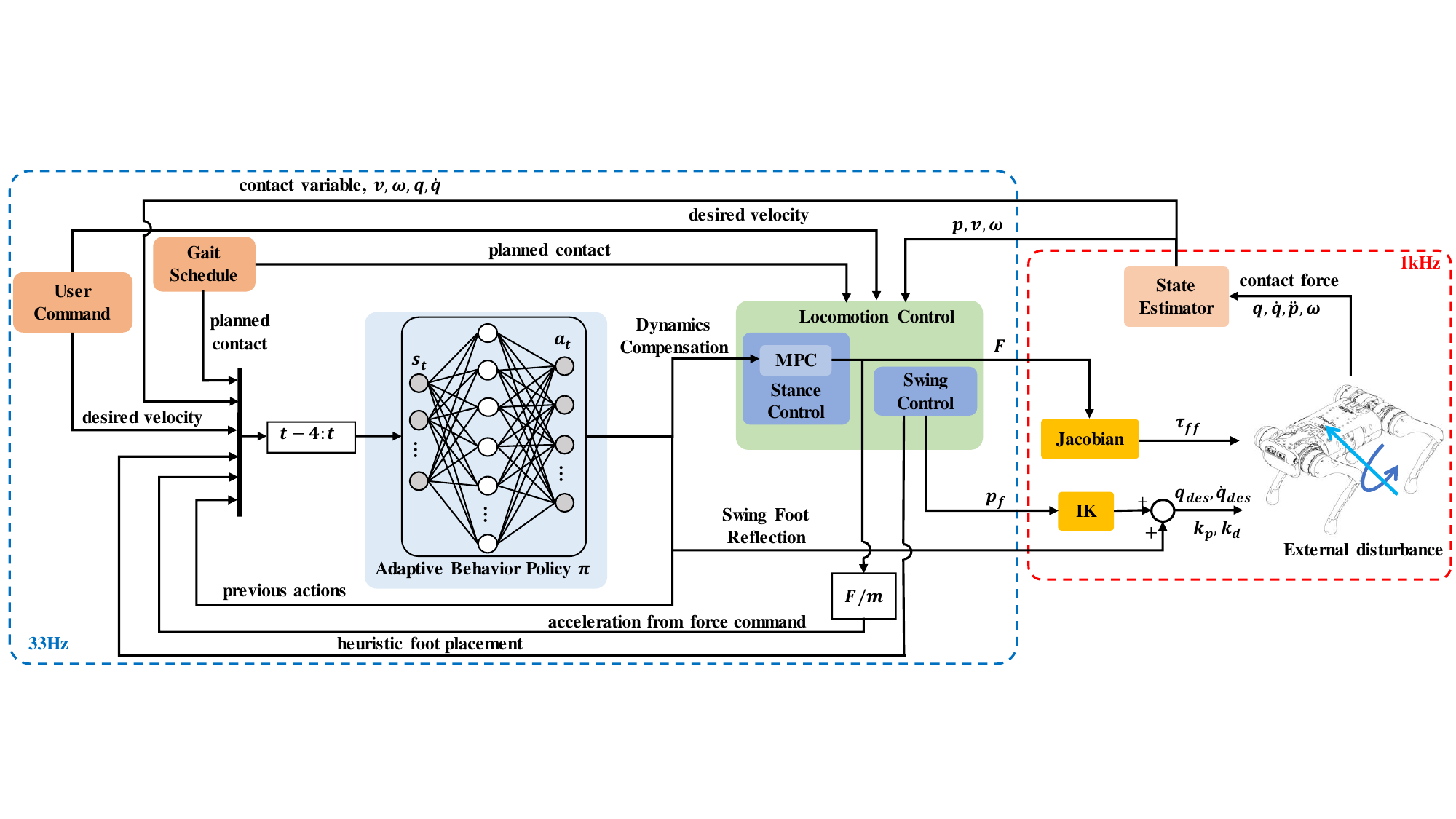}
    \caption{System architecture of the proposed framework. The high-level module, framed in blue, includes the adaptive behavior policy and locomotion control module, operating at 33Hz. The low-level module, running at 1kHz, includes leg control (using Jacobian and IK), state estimation, and the robot's hardware. The $F/m$ block normalizes the MPC force command into accelerations as a robot-agnostic input to the adaptive behavior policy. }
    \label{fig:sys}
    \vspace{-1.5em}
\end{figure*}

\section{Proposed Approach}
\label{sec:approach}
\subsection{System Overview}
Illustrated in \figref{fig:sys} is the overall system architecture, which is built upon \cite{bledt2018cheetah}. The user provides velocity commands to the robot, while an event-driven finite state machine determines the gait schedule. Within the locomotion control module, the MPC is in charge of stance foot control. In contrast, the swing foot control determines the desired foot positions $p_f$. Force commands $F$ are converted into joint torques using the Jacobian, and concurrently, the desired foot positions are mapped to corresponding joint angles through inverse kinematics. A Kalman filter facilitates state estimation, delivering proprioception data to both the locomotion control and the adaptive behavior policy. 

Central to this system is our innovative adaptive behavior policy. Its primary aim is to impose supplementary actions onto the baseline MPC framework to bring adaptive behavior to the robot while ensuring performance across multiple robot platforms. This policy processes past commands, proprioception, acceleration from MPC force commands and heuristic foot placement. The result is dynamic compensation (explained in \secref{sec:mpc_formulation}) for the MPC and an offset joint angle $\Delta q$ for swing foot reflection (explained in \secref{sec:swing}). The adaptive behavior policy (explained in \secref{sec:learning}), which learns to synthesize both the dynamics compensation essential for force control and the reaction for swing foot trajectory, given gait schedule and velocity commands. More than mere compensation, our policy amplifies the agility and robustness of the locomotion. Importantly, it achieves broad generalizability across different robot platforms without resorting to domain randomization. 

\subsection{MPC with Dynamics Compensation}
\label{sec:mpc_formulation}
To address the challenges of model uncertainties while retaining the generalizability across different robot platforms, we build upon the convex MPC setup in \cite{di2018dynamic}. Model uncertainties inherently sprout from: 1) the model mismatch between the simplified model and hardware when abstracting from full-order dynamics, for instance, leg dynamics, and joint-level torque mapping. and 2) the disturbances applied to the robot from unknown disturbances, loads, and terrains.
To capture these uncertainties, we introduce time-varying, locally-linear acceleration terms to incorporate into the linearized continuous-time state space equation. In this way, we incorporate a term in the MPC formulation to encompass different types of model uncertainties. The dynamics compensation terms are encapsulated $\Delta\bm{\alpha}$ and $\Delta\bm{a}$, which represent angular and linear accelerations respectively in the continuous-time state space equation:
\vspace{-0.8em}
\begin{multline}
\label{eq:ct_ss}
 \frac{d}{dt} \left[\begin{array}{c} \bm{\Theta} \\ \bm{p} \\ \bm{\omega} \\ \bm{\dot{p}} \\ \end{array} \right]  = \left[\begin{array}{cccc} 
    \bm{0}_3 & \bm{0}_3 & \bm{R}_z(\psi) & \bm{0}_3 \\ 
    \bm{0}_3 & \bm{0}_3 & \bm{0}_3 & \bm{I}_3 \\ 
    \bm{0}_3 & \bm{0}_3 & \bm{0}_3 & \bm{0}_3 \\ 
    \bm{0}_3 & \bm{0}_3 & \bm{0}_3 & \bm{0}_3\\ \end{array} \right] 
    \left[\begin{array}{c} \bm{\Theta} \\ \bm{p} \\ \bm{\omega} \\ \bm{\dot{p}} \\ \end{array} \right] +\\
    \left[\begin{array}{ccc}
    \bm{0}_3  & ... & \bm{0}_3 \\
    \bm{0}_3  & ... & \bm{0}_3 \\
    \bm{\hat{I}^{-1}}[\bm{r}_1]_{\times} & ... & \bm{\hat{I}^{-1}}[\bm{r}_n]_{\times} \\
    \bm{1}_3/m     & ... & \bm{1}_3/m\\
    \end{array} \right] \left[\begin{array}{c} \bm{F}_0 \\ .\\ . \\ \bm{F}_n \end{array} \right] + 
    \left[\begin{array}{c} 0_{3\times1} \\ 0_{3\times1} \\ \Delta\bm{\alpha} \\ \Delta\bm{a} + \bm{g} \end{array} \right]
\end{multline}
where $\bm{\Theta}$ represents the robot's orientation as a vector of Euler angels $[\phi, \theta, \psi]^{T}$, $R(\psi)$ is the rotation matrix corresponding to the yaw angle $\psi$, $\bm{p}$ and $\bm{\dot{p}}$ is the COM position and velocity of the robot, $\omega$ is the angular velocity of the robot, $r_{i}$ is the vector from the robot's COM to foot $i$, $F_{i}$ is the ground reaction force for leg $i$, $I$ is the inertia, $m$ is the mass of the robot, and $\bm{g}$ is the gravity term.

Equation \eqref{eq:ct_ss} can be rewritten with an auxiliary state to represent the dynamics into a convenient state-space form:
\vspace{-0.5em}
\begin{align}
    \dot{\bm{x}}_c(t) = \bm{A}_c(\psi, \Delta\bm{\alpha}, \Delta\bm{a})\bm{x}_c(t) + \bm{B}_c(\bm{r}_{1...n}, \psi)\bm{u}(t)
\end{align}
where
\vspace{-1em}
\begin{multline}
    \bm{x}_c(t) = \left[\begin{array}{cccccc} \bm{\Theta} & \bm{p} & \bm{\omega} & \bm{\dot{p}} & 1 \end{array}\right]^{T} \in  \mathbb{R}^{13} \\
    \bm{A}_c(t) = \left[\begin{array}{ccccc} 
    \bm{0}_3 & \bm{0}_3 & \bm{R}_z(\psi) & \bm{0}_3 & \bm{0}_{3\times1} \\ 
    \bm{0}_3 & \bm{0}_3 & \bm{0}_3 & \bm{I}_3 & \bm{0}_{3\times1} \\ 
    \bm{0}_3 & \bm{0}_3 & \bm{0}_3 & \bm{0}_3 & \Delta\bm{\alpha} \\ 
    \bm{0}_3 & \bm{0}_3 & \bm{0}_3 & \bm{0}_3 & \Delta\bm{a} + \bm{g} \\ 
    \bm{0}_{1\times3} &  \bm{0}_{1\times3} &  \bm{0}_{1\times3} &  \bm{0}_{1\times3} & 0 \\
    \end{array} \right] \\ \in  \mathbb{R}^{13 \times 13}
\end{multline}
Then, this formulation is discretized and formulated as a QP problem as in \cite{di2018dynamic}.

Opting for accelerations over forces and moments offers a broader view of disturbances to scale the dynamics for different robot platforms. This is particularly important considering robots vary in their ability to withstand external forces and moments. Consequently, we choose a metric that stands independent of the unique mass characteristics specific to each robot, recognizing that these attributes play a pivotal role in adaptability. It's noteworthy that even if the inertia of robots is usually minimal, any oversight in compensating moments can critically impair controller efficacy. Angular acceleration $\bm{\Delta\alpha}$, in this respect, offers a more intuitive and efficient mechanism to modulate the robot's orientation. To delve deeper into this nuance: the process of translating force/moment to acceleration inherently requires knowledge of the robot's mass and inertia. This perspective inherently considers the robot's mass and inertia, allowing our approach to seamlessly generalize across various robotic platforms regardless of mass and inertia differences.  Moreover, this design choice also allows a compact formulation of the optimization problem as the size of all the matrices remains the same as in \cite{di2018dynamic}, which also facilitates onboard computation. 
\vspace{-0.5em}
\subsection{Adaptive Foot Swing Reflection}
\label{sec:swing}
The swing foot in legged robots involves two main components: 1) foot placement and 2) swing trajectory. The foot placement determines the contact location which is crucial for stance control, while the swing trajectory would help the foot reflection to overcome obstacles. Our goal transcends the conventional scope of adaptive control that solely addresses model uncertainties. Instead, we seek to attain adaptive behavior in both the foot placement and swing trajectory, responding reactively to external disturbances, such as external loads or varying terrain.

In the baseline MPC framework, foot placement follows a predetermined heuristic\cite{kim2019highly} based on velocity commands, feedback, and stance time:
\vspace{-0.5em}
\begin{multline}
    \bm{p}_{heuristic,i} = \bm{p}_{hip,i} + \frac{T_{stance}}{2}\bm{v} + k(\bm{v} - \bm{v}_{cmd}) \\
    + \frac{1}{2}\sqrt{\frac{z_{0}}{||g||}}\bm{v} \times \bm{\omega}_{cmd}
\end{multline}

where $\bm{p}_{hip,i}$ is the hip location in the world frame for leg $i$, $T_{stance}$ is the scheduled stance phase time, $\bm{v}$ is the velocity of the robot's COM, $\bm{v_{cmd}}$ is the velocity command, $z_0$ is the nominal height of locomotion, $\bm{\omega}_{cmd}$ is the yaw rate command and in this setup, we used a $k$ of $0.03$. We then employ a pre-defined Bezier curve to interpolate the foot swing trajectory, outputting the $p_f$ as the desired foot location to the swing foot.


Our methodology places a heightened emphasis on adaptive swing reflection. Unlike traditional approaches that manually adjust trajectories, our system utilizes an offset joint angle, $\Delta q$, to modulate the nominal swing trajectory.  This doesn't just modify foot placement; it also dynamically adjusts its swing trajectory over discrete obstacles. By prioritizing adaptive swing control, our system offers a more holistic and responsive solution, synchronously adjusting both the trajectory shape and final foot placement in real time, all under the guidance of one unified adaptive behavior policy. When combined with dynamic compensation, our action space enhances the robustness and agility of legged locomotion.
\vspace{-0.5em}
\subsection{Learning to Synthesize Stance Control and Swing Control for Adaptive Behavior}
\label{sec:learning}
In this section, we present our novel learning framework that integrates the strengths of the traditional MPC control approach with the dynamic adaptability offered by Reinforcement Learning (RL). While the baseline MPC framework excels in forward-looking predictions, RL is capable of reasoning over past experiences. Our primary goal transcends merely addressing model uncertainties; we strive to synthesize these decoupled control realms (stance foot control and swing foot control) using RL. This method unravels the intricate connections between stance foot and swing foot controls. 
This synthesis means that when force optimization is constantly evolving, enriched by insights from the swing foot's heuristic and past proprioception data. In parallel, the swing foot's trajectory and placement are fine-tuned based on cues from the force optimizations and past proprioception data. This seamless integration and reciprocal adaptation ensure that the robot exhibits adaptive behavior under different conditions, underscoring the power and efficacy of our proposed approach.

\subsubsection{Action Space} Considering that the MPC problem intrinsically incorporates the robot's mass properties, we can design robot-agnostic action space to account for model uncertainty while ensuring generalizability. Our learning module computes both dynamics compensation components - namely $[\Delta\bm{\alpha}, \Delta\bm{a}]$ - and swing foot reflection in the form of joint angle offset $\Delta q$ from the nominal swing trajectory. In other words, our adaptive behavior policy seeks to derive sacalable supplementary actions that can be seamlessly layered onto the locomotion controls for different robot platforms. This design ensures improved robustness and agility in locomotion performance and substantially facilitates the generalizability of the framework. 

\subsubsection{Observation Space}
To ensure generalizability, the observation space must remain independent of the robot's mass properties, because the MPC problem inherently considers them. At every time $t$. the policy obtains an observation and performs supplementary actions to the nominal MPC control framework. As presented in \figref{fig:sys}, the observation for the policy takes a history window of 5 MPC horizons, capturing parameters including the joint angle and velocities $\bm{q}$ and $\bm{\dot{q}}$, linear and angular velocities of the robot $\bm{v}_{com}$ and $\bm{\omega}_{com}$, planned contact boolean of every foot from the given gait schedule $\bm{s}_{\phi}$, the actual contact state of each foot from the contact sensor data $\bm{s}_{actual}$, desired COM state from user input $\bm{v}_{des}$ and $\bm{\omega}_{des}$, the heuristic foot placement of every foot $\bm{p}_{heuristic}$ and the acceleration from force commands of MPC optimization $\bm{F}/m$. Similar to dynamics compensation, the ground reaction force command is expressed in terms of acceleration. This consideration is pivotal for generalization across different robots, given that robots come with diverse mass properties, prompting the MPC to produce force commands on varying scales. By ensuring the observation space remains agnostic to a robot's unique internal attributes, we substantially facilitate the generalizability of our framework.


\subsubsection{Training}
In this paper, we employ PPO\cite{schulman2017proximal} for training and use the Unitree A1 robot in the simulation. The reward function at time $t$ is designed to ensure velocity tracking of the robot while minimizing the energy cost.
\begin{align}
    R(t) = w_1 r_{survival} + w_2 r_{velocity} + w_3 r_{energy} + w_4 r_{height}
\end{align}
\vspace{-2em}
where 
  \vspace{2em}
\begin{multline} 
                       r_{survival} = 1 \\
r_{velocity} = ||\bm{v}_{des} - \bm{v}_{COM}|| + ||\bm{\omega}_{des} - \bm{\omega}|| \\
r_{energy} = \sum_{i = 1}^{12} ||\tau_{i}  \dot{q}_{i}||t \\
r_{height} = 0.02 - ||z_{COM} - z_{des}||
\end{multline}

and $w_{i}$ are the corresponding weight factors for each reward term. We use the same reward function by varying weights for the 3 applications we validate our approach. In this work, we approximate the policy using MLPs with hidden layers of [256, 32, 256] neurons with tanh as activation function. The training of the MLP is performed offline with numerical simulation by Pybullet\cite{coumans2021}. 

\subsection{Learning with MPC in Simulation}
In the architectural design of our training methodology, the MPC setup holds a central position. The integration of MPC becomes a computation bottleneck, prominently manifested in the form of optimization-induced latency. With the QPOASES\cite{Ferreau2014} solver, the MPC problem's computation time averages 1$ms$. In comparison, a simulation step in Pybullet is computed in less than 0.1$ms$. The implication here is clear: updating the ground reaction force at each simulation step would drastically decelerate simulation and consequently slow done training. Therefore, our strategy is to update the MPC problem at a less frequent rate, while the lower-level joint commands receive updates much more frequently. As illustrated in \figref{fig:sys}, we update the MPC problem every 30 $ms$ (MPC horizon time) to generate the desired ground reaction force command while the Jacobian maps this force command to joint torques every 1 $ms$. We also have our policy to be set up with an update rate of 33 $Hz$. This configuration aligns with our hardware experiments, negating the need to constantly run the MPC solver for updating the ground reaction force at every control step. Thanks to this efficient setup, our approach not only expedites the training but also ensures the sim2real of the framework as it mirrors the exact setup used in our hardware experiments.
\section{Experimental Validation}
\label{sec:result}

In this section, we detail the experimental validation of our framework, highlighting the enhanced robustness, agility, and adaptive behavior in blind locomotion. Our approach is rigorously tested across various computation platforms and robotic systems, with all policy learning conducted offline using the Unitree A1 robot. For the comparative study, we maintained uniformity by adhering to the baseline MPC specifications, which include gait patterns, foot swing height, MPC weights, and joint PD gains. Throughout the experiments, the robot was maneuvered using a joystick by the author. A seamless transfer of all policies to the actual hardware was made possible thanks to the robustness of the MPC controller and the generalizability of the learning framework.
\vspace{-0.5em}
\subsection{High Speed Locomotion}
The primary aim here is to validate our RL-augmented MPC methodology during challenging high-speed maneuvers—both running and turning. Due to their stark dynamical distinctions, these activities provide a rigorous testbed. We use a flying trot gait for these high-speed maneuvers. The learning domain comprises dynamics compensation factors and foot placement offsets, delineated as The angular acceleration within $\pm[2.0, 10.0, 2.0] rad/s^{2}$, the linear acceleration bounded by $\pm[4.0, 2.0, 3.0] m/s^{2}$ and the joint angle set to $\pm0.3 rad$. 
\subsubsection{High Speed Turning}
\begin{figure}[!t]
    \includegraphics[clip,width=\linewidth]{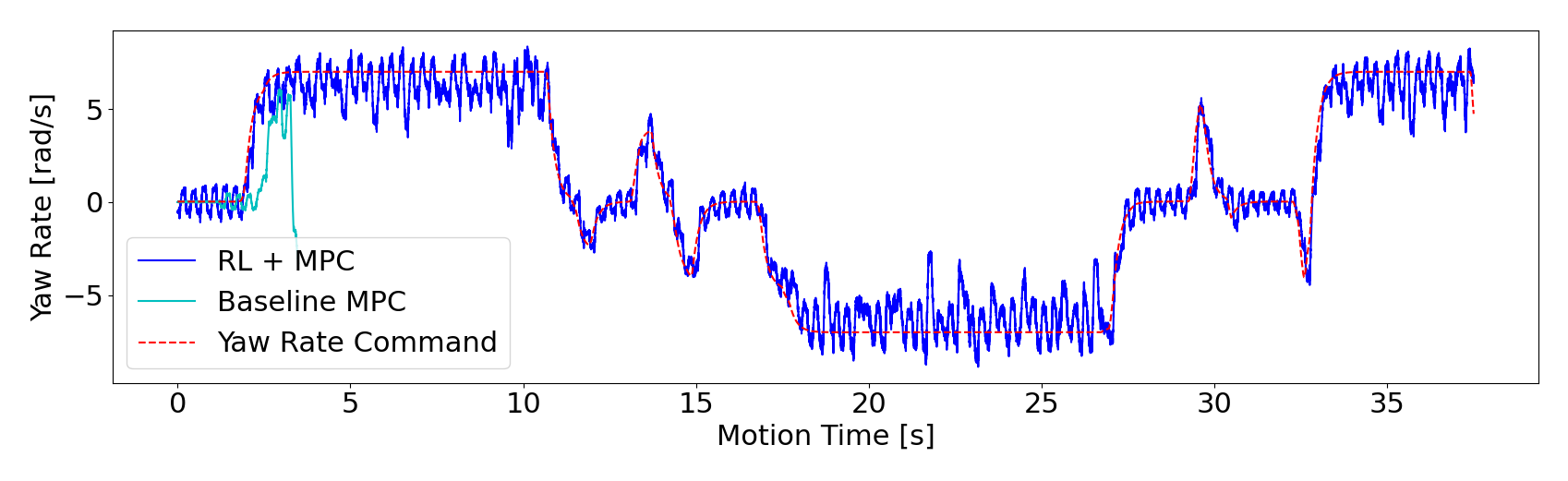}
    \caption{Yaw rate plot of high speed turning policy from IMU data}
    \label{fig:yaw_rate}
\end{figure}

We validated our approach with high-speed turning in place and we are able to achieve state-of-the-art performance in terms of the turning rate. As presented in \figref{fig:yaw_rate}, we ramp up the yaw rate command to $\pm7 rad/s$. Impressively, the A1 robot hit a peak turn rate of 8.5 $rad/s$ in the counter-clockwise direction, maintaining an average yaw rate close to $7 rad/s$, while the baseline MPC couldn't survive this command and failed immediately. As demonstrated in the support video, the policy doesn't merely rely on adjusting foot placement to enhance motion. A significant contribution comes from the angular acceleration compensation in the roll direction as shown in \figref{fig:highlight}. This intelligent adjustment by the policy facilitates smoother turns, enabling us to register the swiftest turn ever recorded on the Unitree A1 robot.

\subsubsection{High-Speed Running and Steering}
\begin{figure}[!t]
\vspace{-1em}
    \centering
    \includegraphics[clip,width=\linewidth]{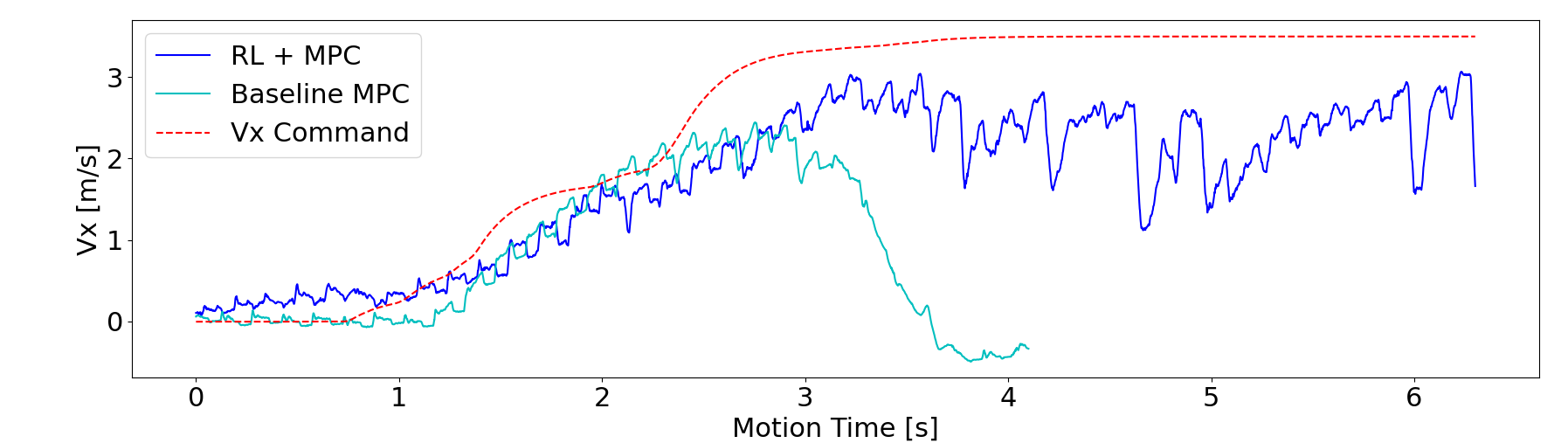}
    \caption{Linear velocity in the body frame of high-speed running policy}
    \label{fig:vel}
    \vspace{-2em}
\end{figure}

\begin{figure}[!tbh]
    \centering
    \includegraphics[clip,width=\linewidth]{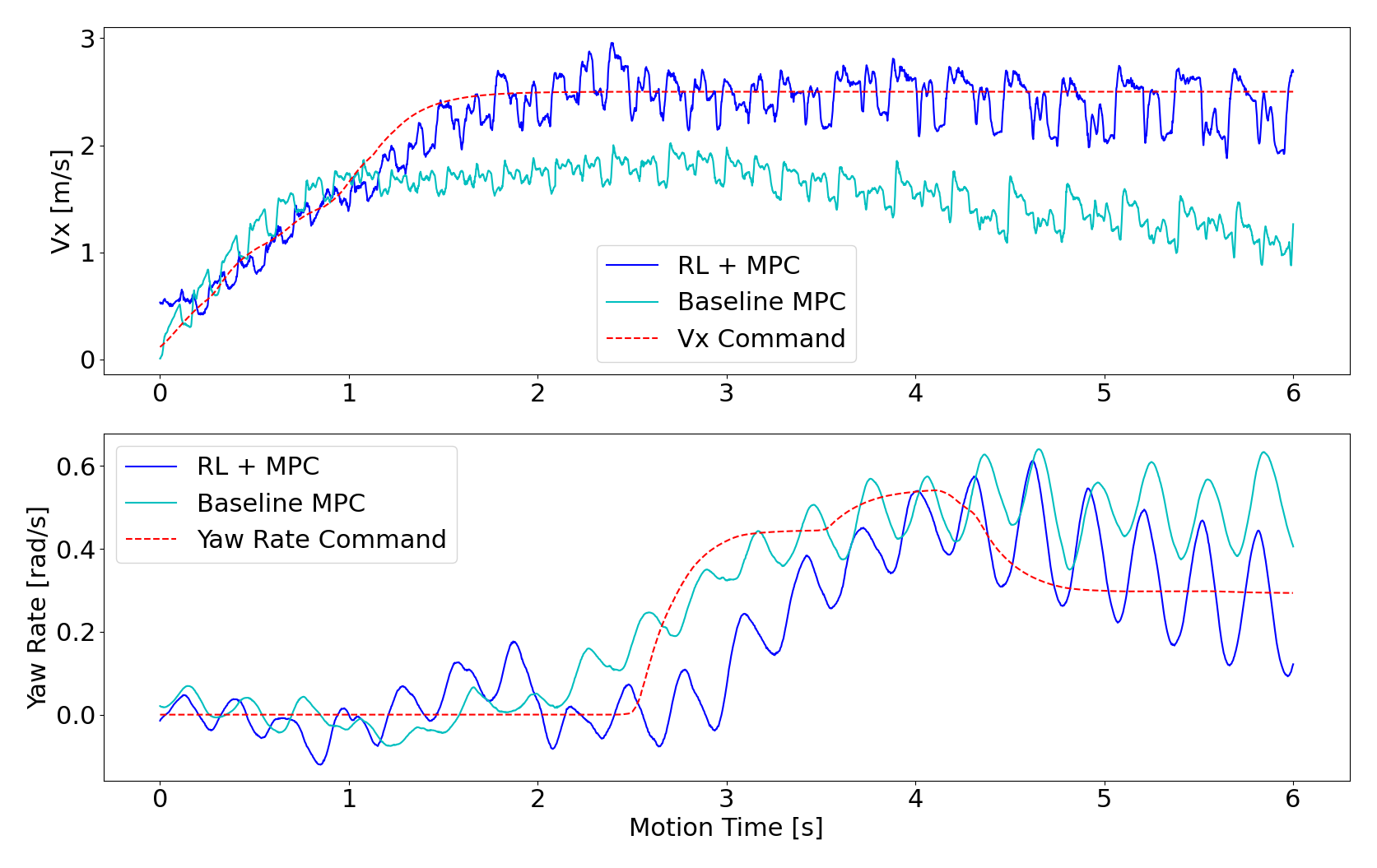}
    \caption{Linear velocity in body frame and yaw rate of high-speed running and steering policy}
    \label{fig:run_turn}
\end{figure}

Furthermore, we conducted tests involving high-speed running and steering. In experiments depicted in \figref{fig:vel}, when velocity commands were increased to $3.5 m/s$, the baseline MPC failed within a few running steps due to the model uncertainty at high speed. Contrarily, our learned policy enabled the robot to withstand disturbances at high speeds, achieving a peak velocity of around $3 m/s$ based on state estimation data. Further tests involved steering at high speed. We first accelerated the robot to $2.5 m/s$ and then applied a yaw rate command of $0.5 rad/s$. \figref{fig:run_turn} suggests that our learned policy surpassed the baseline MPC in velocity tracking. Prior to turning, our RL-augmented MPC  tracked the intended velocity of $2.5m/s$, while the baseline lagged, peaking at roughly $2m/s$. Integrating translation and rotation dynamics, our RL-augmented MPC deftly steered at $2.5m/s$ as highlighted in \figref{fig:highlight}. Meanwhile, the baseline MPC decelerates to $1m/s$ to prevent failure.

\vspace{-0.5em}
\subsection{Walking with Significant Model Uncertainty}
In this section, we highlight our framework's effectiveness in managing model uncertainty and enhancing the robustness of locomotion. In the training setup, we randomize the velocity commands for the robot in body frame of $\bm{v}_x \in{[-1, 1]m/s}$, $\bm{v}_y \in {[-0.5, 0.5]m/s}$ and $\bm{\omega}_z \in{[-2.0, 2.0]rad/s}$ given a trotting gait of 0.3$s$ gait cycle. We choose trotting as the test gait as it is more difficult than standing or quasi-static walking. Adding to this complexity, random external forces and moments are applied, challenging the robot further. Within this setting, agents learn dynamics compensation and foot placement offsets, with angular acceleration constrained to $\pm[4.0, 10.0, 2.0] rad/s^{2}$,  linear acceleration at $\pm[4.0, 2.0, 8.0] m/s^{2}$ and the joint angle set to $\pm0.3 rad$. 

Notably, even though the policy was primarily trained on flat terrain, it showcased remarkable adaptability on soft, uneven soil terrain, when the A1 robot carries an additional 5$kg$ load, as depicted in \figref{fig:highlight}. Moreover, the generalizability of our approach is evident as the policy learned on the A1 robot realizes zero-shot transfer to different robotic platforms like Go1 and AlienGo while adapting effortlessly to different gait cycle timings (see support video). This generalizability is credited to our scalable action space and observation space, which is independent of the robot's internal properties.

\begin{figure}[!tbh]
    \centering
    \begin{subfigure}[Baseline MPC]{
        \includegraphics[width = 0.4\linewidth]{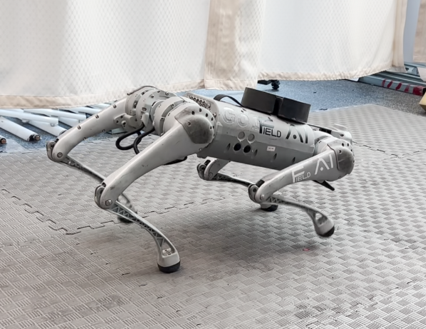}
        \includegraphics[width = 0.4\linewidth]{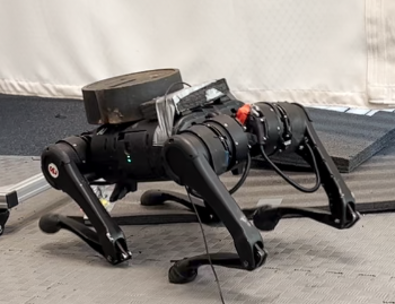}}
    \end{subfigure}
    \vspace{-1em}
    \begin{subfigure}[RL-Augmented MPC]{
        \includegraphics[width = 0.4\linewidth]{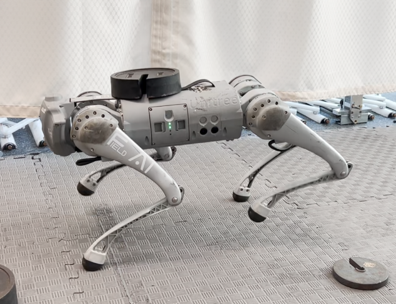}
        \includegraphics[width = 0.4\linewidth]{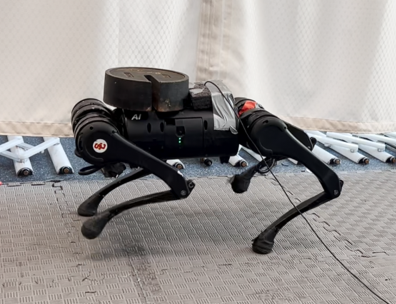}}
    \end{subfigure}
    \caption{Comparison between baseline and proposed approach in terms of pitch compensation. The Go1 robot is carrying a load of 5$kg$ and the A1 robot is carrying a load of 10$kg$. They all run the same adaptive behavior policy trained on the A1 robot.}
     \vspace{-1em}
    \label{fig:pitch_comp}
\end{figure}

Our framework also excels in compensating for external moments. The exemplary adaptive behavior of our framework is evidenced in \figref{fig:pitch_comp}, where we introduce loads that impose additional moments on the robot's body. Despite these challenges, our adaptive behavior policy effectively mitigates the uncertainties. This performance can be attributed to the incorporation of angular acceleration as the key dynamics compensation term. Furthermore, our framework's computational efficiency stands out, comfortably running on a Raspberry Pi 4 board on the Go1 robot. The policy inference and MPC optimization respectively clock in at approximately 1$ms$ and 3$ms$ respectively on the Raspberry Pi 4 board.

\begin{table}[tpb]
\centering
\vspace{0.06in}
\caption{Load carrying capacity of the same policy over different robots with trotting gait}
\begin{tabular}{ |c |c |}
\hline
Robot & Maximum Load Capacity\\
\hline
AlienGo & 10$kg$ \\
\hline
Go1 & 7$kg$ \\
\hline
A1 & 10$kg$ \\

\hline

\end{tabular} \\
\label{tab:load}
\vspace{-2em}
\end{table}

\begin{figure}[!tbh]
    \centering
    \begin{subfigure}[Baseline MPC]{
        \includegraphics[width = 0.45\linewidth]{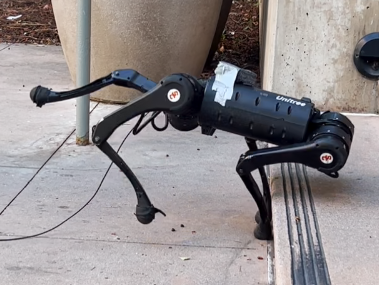}}
    \end{subfigure}
    \begin{subfigure}[RL-Augmented MPC]{
        \includegraphics[width = 0.45\linewidth]{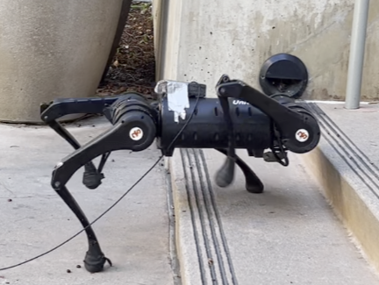}}
    \end{subfigure}
    \vspace{-0.8em}
    \caption{Comparison between baseline and proposed approach for blind stair climbing. The stair height is approximately 13cm and the foot swing height is all set to 8cm.}
    \vspace{-2em}
    \label{fig:stair}
\end{figure}
\vspace{-0.5em}
\subsection{Blind Walking over Discrete Terrain}
We also validated our framework in the realm of adaptive foot swing reflection, specifically on discrete terrain. We trained the policy to traverse over randomly generated discrete terrains ranging from 8 to 12 $cm$ in height, with the robot's foot swing height target fixed at $8 cm$. The action space for this scenario are angular acceleration within $\pm[4.0, 10.0, 2.0] rad/s^{2}$, the linear acceleration at $\pm[2.0, 2.0, 2.0] m/s^{2}$ and the joint angle set to $\pm0.3 rad$. 

In this setup, dynamic compensation handles unexpected ground contact, and adaptive foot behavior prevents foot entrapment. Impressively, this policy seamlessly transitioned to blind stair climbing of a stair height of $13cm$. As presented in \figref{fig:stair}, while the standard MPC often led to the robot's foot being trapped, our refined framework adopted adaptive swing trajectories, facilitating immediate foot reaction upon contact with the discrete obstacle(see support video). This emergent behavior is learned by the policy, and the MPC ensures the robot's stability and movement. The seamless integration is a testament to the effectiveness of our RL-augmented MPC approach for adaptive behavior.

\vspace{-0.5em}
\section{Conclusion}
\label{sec:conclusion}
In this research, we have presented an integration of Reinforcement Learning (RL) and Model Predictive Control (MPC), improving the agility and robustness of legged locomotion. Through the integration of MPC's forward-looking predictions and RL's ability to reason on past experiences, our framework has unveiled marked improvements in a robot's dexterity to traverse intricate landscapes and handle unexpected perturbations. Notably, the adoption of adaptive swing foot reflection showcases how the blend of these two methodologies can lead to real-world locomotion improvements. The extensive experimental validations presented further underscore the robustness and generalizability of our proposed RL-augmented MPC framework, setting a new benchmark for state-of-the-art robotic control systems. We envision that this synthesized approach will pave the way for future research such as perceptive locomotion, fostering even more resilient and adaptive behavior for legged robots. 

\newpage
\bibliographystyle{IEEEtran}
\bibliography{ref}

\end{document}